\documentclass[conference]{IEEEtran}
\IEEEoverridecommandlockouts
\renewcommand{\thesubsection}{\thesection.\alph{subsection}}
\usepackage{tabularx}
\usepackage{cite}
\usepackage{amsmath,amssymb,amsfonts}
\usepackage{algorithmic}
\usepackage{graphicx}
\usepackage{textcomp}
\usepackage{xcolor}
\usepackage{enumitem}
\usepackage{float}
\usepackage{booktabs}
\usepackage{subcaption}
\usepackage{caption}
\begin{document}
\title{Computer Vision and Normalizing Flow Based Defect Detection}

\author{\IEEEauthorblockN{Zijian Kuang\IEEEauthorrefmark{1},
Xinran Tie\IEEEauthorrefmark{2}, Lihang Ying\IEEEauthorrefmark{3}, Shi Jin\IEEEauthorrefmark{4}}
\IEEEauthorblockA{Department of Computing Science\\
University of Alberta\\
Edmonton, Canada\\
Email: \IEEEauthorrefmark{1}kuang@ualberta.ca,
\IEEEauthorrefmark{2}xtie@ualberta.ca,
\IEEEauthorrefmark{3}leo@zerobox.ai,
\IEEEauthorrefmark{4}shi@zerobox.ai}}

\maketitle
\thispagestyle{plain}
\pagestyle{plain}

\begin{abstract}
Visual defect detection is critical to ensure the quality of most products. However, the majority of small and medium-sized manufacturing enterprises still rely on tedious and error-prone human manual inspection. The main reasons include: 1) the existing automated visual defect detection systems require altering production assembly lines, which is time consuming and expensive 2) the existing systems require manually collecting defective samples and labeling them for a comparison-based algorithm or training a machine learning model. This introduces a heavy burden for small and medium-sized manufacturing enterprises as defects do not happen often and are difficult and time-consuming to collect. Furthermore, we cannot exhaustively collect or define all defect types as any new deviation from acceptable products are defects. In this paper, we overcome these challenges and design a three-stage plug-and-play fully automated unsupervised 360-degree defect detection system. In our system, products are freely placed on an unaltered assembly line and receive 360 degree visual inspection with multiple cameras from different angles. As such, the images collected from real-world product assembly lines contain lots of background noise. The products face different angles. The product sizes vary due to the distance to cameras. All these make defect detection much more difficult. Our system use object detection, background subtraction and unsupervised normalizing flow-based defect detection techniques to tackle these difficulties. Experiments show our system can achieve 0.90 AUROC in a real-world non-altered drinkware production assembly line. 
\end{abstract}


\IEEEpeerreviewmaketitle

\section{Introduction}
\begin{figure}[h]
\centering{\includegraphics[width=0.9\columnwidth, height=6cm]{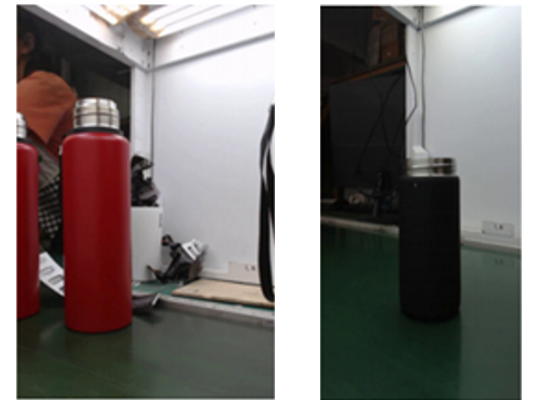}}
    \caption{Examples of data collected from a real-world bottle manufacturer. It demonstrates the complexity and unpredictability of image background noise that could happen in a small to medium sized factory.}
\label{bgnoise}
\end{figure}

Visual defects have a significant impact on the quality of industrial products. Small defects need to be carefully and reliably detected during the process of quality assurance \cite{bottle} \cite{defect1}. It is important to ensure the defective products are identified at earlier stages, which prevents a negative impact on a company’s waste, reputation and additional financial loss. In recent research, visual defect detection has been increasingly studied again with deep learning approaches and has improved quality control in the industrial field \cite{anomaly_survey1} \cite {anomaly_survey2}. However, visual defect detection is still challenging due to 1) collecting defective samples and manually labeling for training is time-consuming; 2) the defects' characteristics are difficult to define as new types of defects can happen any time; 3) and the product videos or images collected from SME's non-altered assembly lines usually contain lots of background noise as shown in Fig. \ref{bgnoise}, since a well designed production lines that can ensure high quality product videos or images can be prohibitively costly for SMEs. The results of defect detection become less reliable because of these factors. 

\begin{figure*}[h]
\centerline{\includegraphics[width=0.9\textwidth, height=8cm]{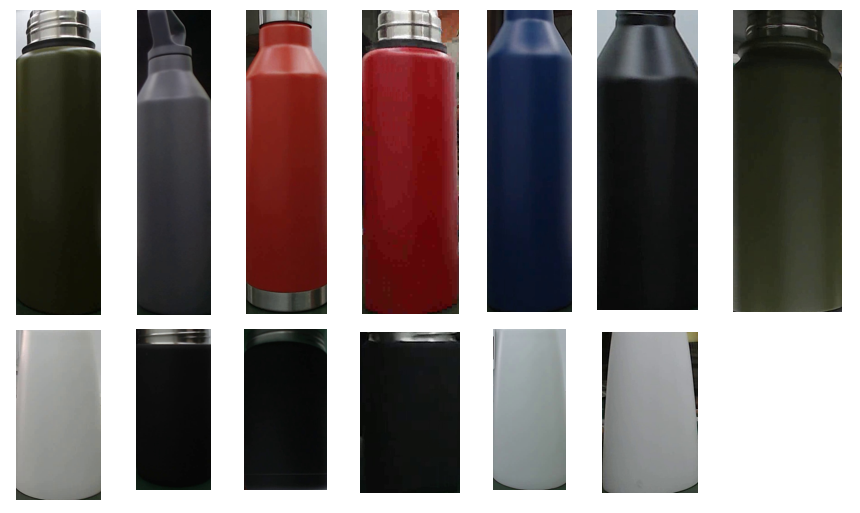}}
    \caption{Samples of the ZeroBox bottle product dataset}
\label{bottle}
\end{figure*}

Most existing defect datasets \cite{mvtec} are either for one scenario (e.g. concrete, textile, etc.) or lack of defect richness and data scale. The popular anomaly defection dataset \cite{mvtec} is too "perfect" (e.g. all products are perfectly aligned in the center of the image, with clean and simple background) which cannot represent the realistic setup in SME factories or requires challenging perfect pre-processing (e.g. background removal, re-lighting, etc). Specifically, the dataset is limited to a few categories of products and a smaller number of samples \cite{bottle}  \cite{defect1} \cite{defect2}. To ensure our experiments' realism and applicability, we introduce a new dataset collected from a commercially operating bottle manufacturer located in China. This dataset includes 21 video clips (with 1634 frames) consisting of multiple types of bottle products with both good and defective samples. Some of them are shown in Fig. \ref{bottle}. These videos are provided by ZeroBox.

Since specialized cameras and well-designed turing assembling lines are too expensive for SME factories, it is highly desirable to have a fully automated defect detection system with minimal cost that can be plug-and-play added to the existing production lines. In this paper, we propose a three-stage deep learning powered, fully automated defect detection system based on object detection, background subtraction and normalizing flow-based defect detection. The system we proposed uses three novel strategies: 

\begin{enumerate}
\item{first, a novel object detection is used to narrow down the searching window and realign the product from each input video frames}
\item{a novel video matting based background subtraction method is used to remove the background of the detected image so that the defect detection model can focus on the product}
\item{finally, a semi-supervised normalizing flow-based model is used to perform product defect detection}
\end{enumerate}

Extensive experiments are conducted on a new dataset collected from the real-world factory production line. We demonstrate that our proposed system can learn on a small number of defect-free samples of single product type. The dataset will also be made public to encourage further studies and research in visual defect detection.

\section{Related Work}
\label{RW}
Since this paper focus on an end to end three stage network for product defect detection, in this section, we will focus on the three areas of object detection, background subtraction and visual defect detection.

\subsection{Object Detection}

Object detection refers to the operation of locating the presence of objects with bounding boxes \cite {object} \cite{yolo-compact}. The types or classes of the located objects in an image are classified by the model with respect to the background. Currently, deep learning-based models are state-of-the-art on the problem of object detection. Top detection frameworks include systems such as deformable parts models, Faster R-CNN, and YOLO.

Deformable part models (DPM) \cite{dpm} use a disjoint pipeline with a sliding window approach to detect objects. The system is disparate and only the static features are extracted. Faster R-CNN \cite{r_cnn} and its variants utilize region proposals to find objects. The pipeline of Faster R-CNN consists of a convolutional neural network, an SVM, and a linear model. However, each of the stages needs to be finetuned precisely and independently. It can not be applied to  real-time situations due to the slowness of the overall system. 

In 2016, J. Redmon et al. introduced a unified real-time object detection model called "You only look once" (YOLO). Unlike DPM and Faster R-CNN, YOLO replaces disparate parts to a single convolutional neural network. It reframes object detection as a regression problem that separates bounding boxes spatially and associates them with their class probabilities \cite{yolo}. YOLO is extremely fast, reasons globally, and learns a more generalized representation of the objects. It achieves efficient performance in both fetching images from the camera and displaying the detections. However, YOLO struggles with small items that appear in groups under strong spatial constraints. It also struggles to identify objects in new or unusual configurations from data it has not seen during the training \cite{yolo}. Still, YOLO is so far the best objection detection algorithm.

\subsection{Background Subtraction}

Background subtraction is a technique that is widely used for detecting moving objects in videos from static cameras and eliminating the background from an image. A foreground mask is generated as the output, which is a binary image containing the pixels belonging to the moving objects \cite {background} \cite{background2}. The methods of background subtraction for videos include video segmentation and video matting. 

In video segmentation, pixels are clustered into two visual layers of foreground and background. In 2015, U-Net \cite{unet} was proposed for solving the problem of biomedical image segmentation. The architecture of this network is in the shape of a letter "U", which contains a contracting path and an expansive path. A usual contracting layer is supplemented with successive layers and max-pooling layers. The other path is a symmetric expanding path that is used to assemble more precise localization. However, excessive data argumentation needs to be applied to retain a considerable size of features if there is a small amount of available training data.

Video matting, as another method of background subtraction, separates the video into two or more layers such as foreground, background and alpha mattes. Unlike video segmentation which generates a binary image by labelling the foreground and background pixels, the matting method also handles those pixels that may belong to both the foreground and background, called the mixed pixel \cite {background} \cite{background2}. Recently, Background Matting V2 (BGM V2) has achieved the state-of-art performance to replace the background in a real-time manner \cite{bgmv2}. The first version of Background Matting (BGM) was initially proposed to create a matte which is the per-pixel foreground colour and alpha of a person in 2020 \cite{bgm}. It only requires an additional photo of the background that is taken without the human subject. Later, Background Matting V2 (BGM V2) is released to achieve real-time, high-resolution background replacement for video conferencing. However, in the final matting results, there is still some residual from the original background shown in the close-ups of users’ hairs and glasses.

\subsection{Defect Detection}

 In recent years, convolutional neural networks began to be applied more often to visual-defect classification problems in industrial and medical image processing. The segmentation approach plays a significant role in visualized data's anomaly detection and localization since it can not only detect defective products but also identify the anomaly area.
 
 Autoencoder has become a popular approach for unsupervised defect segmentation of images. In 2019, P. Bergmann et al. proposed a model to utilize the structural similarity (SSIM) metric with an autoencoder to capture the inter-dependencies between local regions of an image. This model is trained exclusively with defect-free images and able to segment defective regions in an image after training \cite{autoencoder_ssim}. 

Although segmentation-based methods are very intuitive and interpretable, their performance is limited by the fact that Autoencoder can not always yield good reconstruction results for anomalous images. In comparison, the density estimation-based methods can perform anomaly detection with more promising results.

The objective of density estimation is to learn the underlying probability density from a set of independent and identically distributed sample data \cite{density}. In 2020, M. Rudolph et al.\cite{differnet} proposed a normalizing flow-based model called DifferNet, which utilizes a latent space of normalizing flow to represent normal samples' feature distribution. Unlike other generative models such as variational autoencoder (VAE) and GANs, the flow-based generator assigns the bijective mapping between feature space and latent space to a likelihood. Thus a scoring function can be derived to decide if an image contains an anomaly or not. As a result, most common samples will have a high likelihood, while uncommon images will have a lower likelihood. Since DifferNet only requires good product images as the training dataset, defects are not present during training. Therefore, the defective products will be assigned to a lower likelihood, which the scoring function can easily detect the anomalies \cite{differnet}.

\section{Proposed System}
\label{PM}

\begin{figure*}[h]
    \centerline{\includegraphics[width=1.1\textwidth, height=4cm]{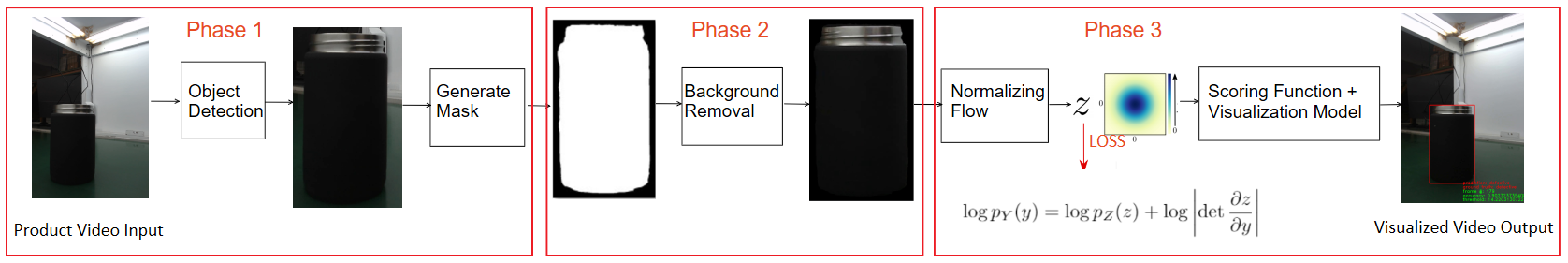}}{}
    \caption{Overview of our proposed system. Phase 1: our system first takes video clips as input and utilizes YOLO to detect and draw bounding boxes on each product in each frame. Phase 2: after YOLO detection, a pretrained background matting model is applied along with our novel background subtraction algorithm to remove the background noises surrounding the product within the bounding box. Phase 3: the processed product images are further passed into the flow-based defect detection model to generate a normal distribution. After training the model, a scoring function is used to calculate likelihoods against the good sample’s distribution to classify the input sample as defective or normal. We also created a visualization model to generate a video output with the bounding box and predicted label on each frame.
}
    \label{fig1}
\end{figure*}
In this paper, we propose a low cost plug-and-play fully automated 360-degree deep learning defect detection system. Without requiring any major re-design of the existing production line, the system is  a simple add-on "box" to the existing process. It utilizes multiple low-cost cameras to capture the product images from different angles to ensure all important visual areas are covered at least once. Then the captured images are used as the input in our proposed deep learning based system to perform defect detection. The overview of the proposed system's pipeline is shown in Fig.\ref{fig1}. 

The general stages and tasks within our proposed product defect detection system can be divided into three main components, which are the object detection (section III.A), the background subtraction (section III.B) and the defect detection (section III.C).

\subsection{Novel object detection based on deep learning and traditional computer vision algorithms}

Our system takes videos of products captured by four cameras installed on the assembling line as input. These cameras are arranged 90 degrees apart around a center spot where all products will pass through along the assembling line.   The 4 camera inputs are fed into the system independently so there is no complication of synchronizing all cameras.

In the video input, the product is moving on the convey belt viewed by a static camera. Therefore the position of the product in each frame is different. In our defect detection model, we want to focus on the product, and to eliminate the unnecessary information from each frame (such as background), we decided to adopt a pre-trained YOLOv5 \cite{yolo} object detection model to narrow down the defect detection searching window on input images collected from each cameras.  The pre-trained YOLOv5 model was further fine-tuned with the ZeroBox dataset.

Even though YOLOv5 is able to detect product position for each frame of the video input, it is computationally too slow to continuously use YOLOv5 for all videos frames from all 4 cameras on a modest computer without GPU.  In order to reduce the computational workload, a traditional computer vision based motion detection algorithm \cite{opencv} is utilized to first identify when a product has moved into the center of each camera view on the conveyor belt and then YOLOv5 is utilized only once per object instead of on all frames of the video stream.

At the end of the object detection stage, the product will be realigned into the center of the bounding box,and around 80\% background information will be eliminated from the original input frames.

\subsection{Novel background subtraction based on video matting and traditional computer vision algorithms}

At the end of the first stage, most of the background has been removed by the YOLOv5 algorithm. However, as depicted by Fig-\ref{fig1}, YOLOv5 would still keep a small margin around the product itself. The background in the margin makes it  difficult for defect detection algorithms since background can vary significantly from image to image and is often mistreated as defective. Since YOLOv5 is only able to identify objects by rectangular boxes, this problem is  particularly challenging for products that don't fit snugly in the bounding box such as some of the products shown in Fig-\ref{bottle} (for example the cone shaped bottles and those with smaller necks).

To overcome this problem, an image background subtraction model is further utilized to remove the background in each YOLOv5 bounding box. After object detection and background subtraction, the processed images will  be 100\% of the product itself  and then suitable to be passed on to defect detection phase.

\begin{figure*}[h]
\centerline{\includegraphics[width=0.8\textwidth]{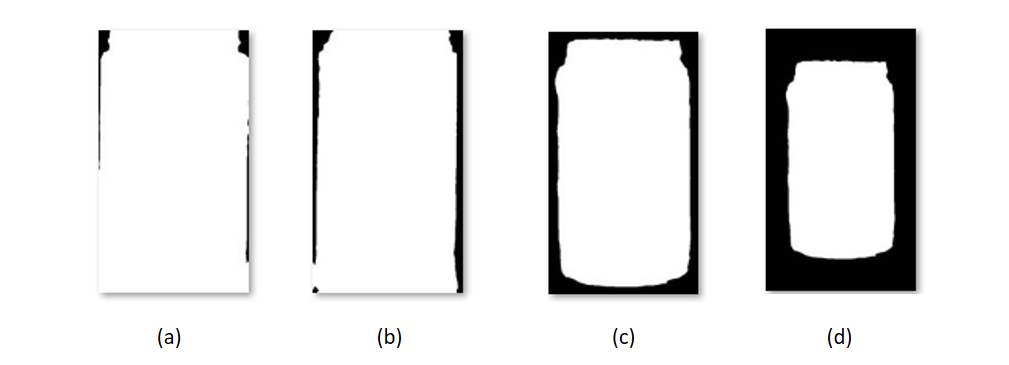}}
\caption{Examples of mask generated in different stages. (a) Mask generated using BGMV2 on first frame of the black bottle product. (b) Composite first 10 masks generated using BGMV2 on first 10 frames. (c) The composite mask generated using the entire video dataset. (d) The 10\% shrink and padding-resize of composite mask to minimize the background from each frame}
\label{mask}
\end{figure*}

We use the background matting technique from BGMv2 \cite{bgmv2} to draw a mask to remove the background. However, the matting performance is not very reliable. The mask generated in each frame is slightly different from the mask generated in other frames. To overcome this issue, we propose to linearly add the masks that are generated in all bounding boxes from sequential video frames as a composite mask. In other words, in the single product video input, we will generate one single mask to segment the product and background in every single frame. Then we use the composite mask to remove the background from each bounding box generated by YOLOv5 in each frame.

\begin{figure*}[h]
\centerline{\includegraphics[width=0.9\textwidth]{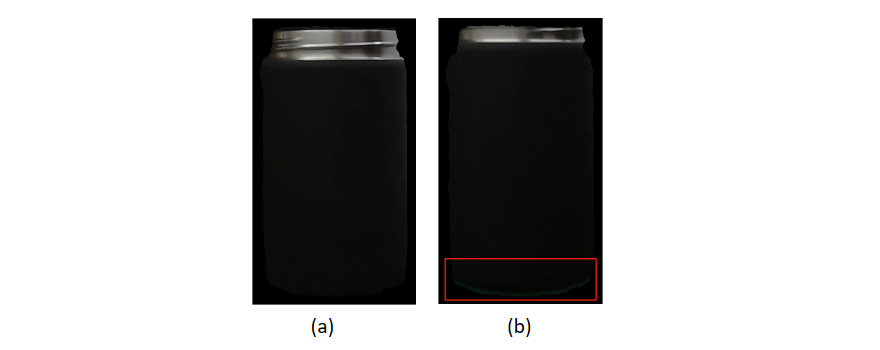}}
\caption{Examples of results after background removal using composite masks. (a) Ideal background subtraction using composite mask (b) Some cases that the composite mask cannot remove all the background due to the product might not be in the center of the image}
\label{bgm}
\end{figure*}

Since every frame is different, the generated composite mask cannot always fully remove all the background. As shown in Fig. \ref{bgm}(b), the bottom of the image still include some conveyor belt portion which is considered as the background noise. To solve this problem, we further shrink the final mask by a fixed percentage  and the mask is then padded to the original size as shown in Fig. \ref{mask}(d). The re-scaled mask can ensure all the background are removed in every frame. The areas near the boundaries of the product can also be masked, so our defect detection model might miss the defects in these boundary regions. However, this problem is compensated by the fact multiple cameras are employed in the system: most defects missed in one camera near the product boundary is fully visible close to the center view of another camera.

\subsection{Defect detection using normalizing flow based model}
After object detection and background subtraction, the processed images are further resized to the size of 448 by 448 pixels that only contains the product information excluding any background noise. Then the processed images are fed into DifferNet \cite{differnet} to output a normal distribution by maximum likelihood training. To classify if an input image is anomalous or not, our model uses a scoring function that calculates the average of the negative log-likelihoods using multiple transformations of an image. The result will compare with the threshold value which is learned during training and validation process and is later applied to detect if the image contains an anomaly or not \cite{differnet}. More implementation details and threshold selection strategy along with experiment results are shown in the next section.

After defect detection, the information include anomaly prediction and predicted bounding box will be plot onto the original product video input as our visualized video output. The example frame of output result can be found in Fig. \ref{fig1}. Since we have four cameras to capture the 360-degree images of the product, the product will be classified as defective if any of the cameras detects a defect.

\section{Experiments and Results}
In this section, we evaluate the proposed system based on real-world videos provided by a factory in China. First, we briefly introduce the dataset that is used in the following experiments. Then, the results of several representative experiments are studied along with visual statistics. Since the complexity of experiments primarily stems from the noisy background in the video inputs, our experiments will solely concentrate on logo-free products and group them into single and multiple product categories for experiment purposes.

\subsection{Dataset}

\begin{figure*}[h]
    \centering
    \begin{subfigure}[b]{0.4\linewidth}
        \centering
        \includegraphics[width=\textwidth, height = 7cm]{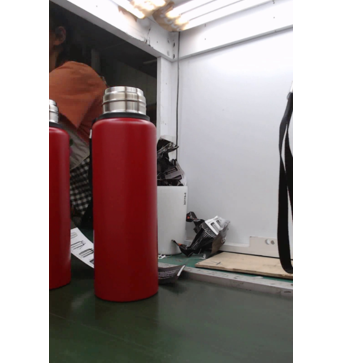}
        \caption{}
        \label{fig:18}
    \end{subfigure}
    \begin{subfigure}[b]{0.4\linewidth}  
        \centering 
        \includegraphics[width=\linewidth, height = 7cm]{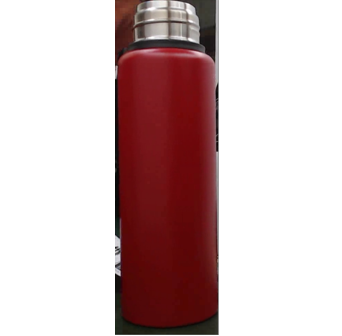}
        \caption{}
        \label{fig:19}
    \end{subfigure}
    \begin{subfigure}[b]{0.4\linewidth}   
        \centering 
        \includegraphics[width=\linewidth, height = 7cm]{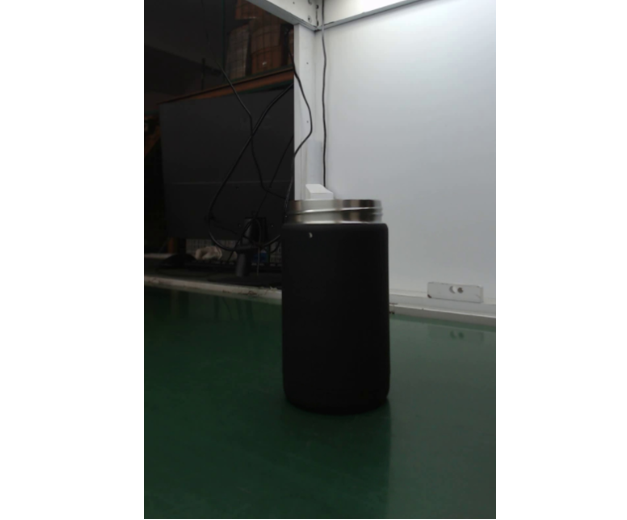}
        \caption{}  
        \label{fig:20}
    \end{subfigure}
    \begin{subfigure}[b]{0.4\linewidth}   
        \centering 
        \includegraphics[width=\linewidth, height = 7cm]{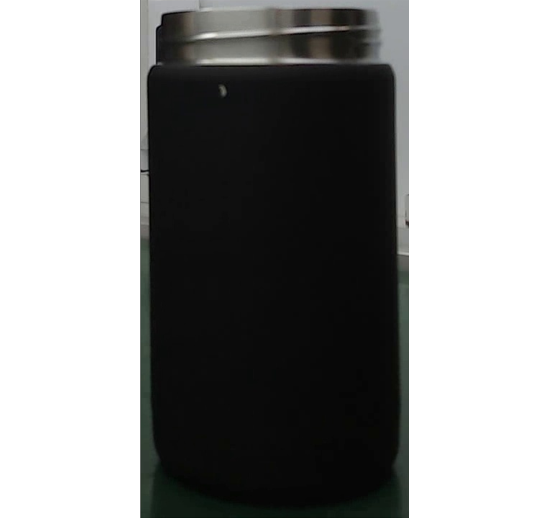}
        \caption{}
        \label{fig:21}
    \end{subfigure}
    \caption{ Example images from the ZeroBox dataset of products from a real-world bottle factory. Fig. \ref{fig:18} and Fig. \ref{fig:19} show examples of the original image and cropped images of a good product. Fig. \ref{fig:20} and Fig. \ref{fig:21} show examples of the original  and cropped images of a defective product.}
    \label{fig:22}
\end{figure*}

In this paper, we evaluate our defect detection system based on videos recorded from real life. For fair and reliable experiment results, ZeroBox Inc. has created a brand new dataset collected from an industrial production line monitoring system. This dataset includes 21 video clips in total which consists of 13 types of products with both good and defective samples. Some of the product samples are shown in Fig. \ref{bottle}.  In addition, there are 1381 good product images and 253 defective product images generated through YOLO detection and cropping. Examples of defective and defective-free samples are presented in Fig. \ref{fig:22}. 

Since our normalizing flow-based defect detection model utilizes semi-supervised learning, it only requires approximately 150 good sample images to efficiently learn how to use a simple normal distribution to represent the complex distribution of a group of good samples. Within our experiments on the product of a white jar and a black jar, a total of 150 good sample images are used for training. Another 121 sample images are used for validation, and the rest of 47 sample images are used for testing purposes. For both validation and testing, a mixture of good and defective samples is used to evaluate the proposed system.

\subsection{Implementation Details}
For all experiments, we train our system for 10 meta epochs. Each meta epoch contains 8 sub epochs which result in a sum of 80 epochs. Additional transformation settings are applied to manipulate and adjust brightness, contrast and saturation for the input images. During our experiments, we manipulated the contrast, brightness and saturation of the product images with a uniformly distributed random factor in the interval of $[0.5, 1.5]$ to pre-process input video frames. Results in later sections also compare each experiment outcome while including or excluding this process from the experiment. Although the model DifferNet proposed in the paper \cite{differnet} does not need defective samples during training, this process is still necessary for validation. In fact, the validation process plays a critical role in determining the threshold of the anomaly score. Within the evaluation stage, our proposed system will be validated once at the end of each epoch based on the anomaly score calculated from the current training stage.

During the testing stage, the threshold for detection is chosen based on the corresponding true-positive rate and false-positive rate of the trained model and a given true-positive rate as the target used for training purposes. More specifically, the threshold value has the true-positive rate greater than our target-true positive rate but the smallest false-positive rate will finally be chosen in the testing process to reflect the performance of our system. In order to predict if a given input sample has a good or a defective product in it, we will use the aforementioned threshold to evaluate. If the predicted anomaly score provided by the model is less than the threshold, this sample will be classified as good. Otherwise, the sample is classified as a defective one. Within each meta epoch, the corresponding Area Under Receiver Operator Characteristics (AUROC) along with the calculated threshold values and anomaly scores on our validation dataset for the system are computed and saved for later evaluation stages. The metric AUROC is calculated using build-in \textit{roc\_curve} function imported from scikit-learn library \cite{sklearn}. In the last meta epoch, the system’s aforementioned parameters are saved into the system as well for evaluation. After training and evaluation, the test accuracy is calculated based on the percentage of the test dataset that is correctly classified. Moreover, a ROC curve is plotted at the end of the training and testing process and is saved locally for further analysis and exploration.

\subsection{Detection}
Our proposed system is tested on 2 products; one is jars with black visual and the other is jars with white visual to compare and report the performance. The two jars are in a shape that is very similar to a cone with no logo on their surfaces. Moreover, we have compared the performance of the system trained on 3 different types of input images: original product images, product images with cropping, and product images with a mask used for background removal. From our experiments, the best performance occurs on the input images using the strategy of mask for background removal with a target true positive rate set to $0.85$ in training. Since many of the defects happen far from the edges of the product from the input frames, an extra 10\% mask extension can further enhance the performance and achieve a promising test accuracy. As a result, the accuracy of defect detection is increased by 20\% with a final test accuracy above 80\% as our best performance from the proposed system. Later experiments also show that the effect of background factors in each frame can be further reduced by extending the mask. Detailed experiment results are displayed in the following sections. \\

Experiment results are displayed in detection accuracy and its calculated threshold for each trial and are compared for each of the aforementioned strategies in the tables below. Table \ref{tab:table1} present the detailed performance of detection in terms of test accuracy and its corresponding anomaly threshold on the black jar product. Experiments with and without the image transformation process (manipulation of contrast, brightness and saturation settings of input images) are performed for comparison. Three experiment settings are applied for experiments which include using original images, using images with cropping technique or using images with a mask. By using original images of the product in detection, the proposed system achieves the same test accuracy before and after adding the transformation process. With a 10\% cropping on each side of the image, the proposed system can obtain a better result on defect detection with cropping performed on each side of input images during training. Finally, with an adaptive mask applied to the input images, the proposed system can obtain the best result of 87.00\% as the test accuracy in all the experiments and a value of 55.11 as the corresponding anomaly threshold on defect detection while the mask helps sufficiently eliminate other factors that potentially affect the accuracy in performance.
            
            \begin{table}[h]
            \captionsetup{font=small}
            \centering
            \begin{tabularx}{0.95\columnwidth}{lXX}
            \toprule
             Accuracy/Threshold & Without Transformations & With Transformations \\
            \midrule
            Original Images           & 24.19\% / 4.55    & 24.19\% / 15.88                  \\
            Images with 10\% Cropping & 61.29\% / 84.5 & 75.81\% / 13.91                  \\
            Images with Mask          & 67.42\% / 12.79   & \textbf{87.00\% / 55.11} \\
            \bottomrule
            \end{tabularx}
            \caption{Performance of Detection on Images of Black Jar Represented in Accuracy / Threshold}
            \label{tab:table1}
            \end{table}

Table \ref{tab:table2} present the detailed performance of detection in test accuracy and its corresponding threshold on the product of white jar under the same test setup. Test results are compared with and without the image transformation process for each proposed strategy. In this case, the proposed system has achieved a generally higher test accuracy on images under image transformation settings. The best performance still happens in the case which utilizes the adaptive mask to eliminate the impact from the background.
        
            \begin{table}[h!]
            \captionsetup{font=small}
            \centering
            \begin{tabularx}{0.95\columnwidth}{lXX}
            \toprule
            Accuracy/Threshold & Without Transformations & With Transformations \\
            \midrule
            Original Images           & 24.87\% / 1.71  & 29.09\% / 1.85                   \\
            Images with 10\% Cropping & 65.88\% / 35.76 & 74.11\% / 3.79                   \\
            Images with Mask          & 67.27\% / 8.89  & \textbf{83.63\% / 8.76}     \\     
            \bottomrule
            \end{tabularx}
            \caption{Performance of Detection on Images of White Jar Represented in Accuracy / Threshold}
            \label{tab:table2}
            \end{table}

\section{Conclusion}
In this paper, we introduce a new dataset for product visual defect detection. This dataset has several challenges regarding defect types, background noise, and dataset sizes. We have proposed a three-stage defect detection system that is based on the techniques of object detection, background subtraction and normalizing flow-based defect detection. Finally, extensive experiments show that the proposed approach is robust for the detection of visual defects on real-world product videos. In the future, we plan to work on using background and foreground segmentation with an end-to-end trained mask to eliminate the background noise in images identified by YOLO. Also, more data samples will be collected for training, validation and testing.

\bibliographystyle{IEEEtran}
\bibliography{main}

\end{document}